\documentclass{llncs}
\usepackage{makeidx}  
\usepackage{comment}
\usepackage[colorinlistoftodos]{todonotes}
\usepackage{graphicx}
\usepackage[compatibility=false]{caption}
\usepackage{subcaption}

\begin{document}
\frontmatter          
\pagestyle{headings}  

\mainmatter              
\title{Introspective Agents: Confidence Measures for General Value Functions}
\titlerunning{Introspective Agents: Confidence Measures for General Value Functions}  
%
\author{Craig Sherstan\inst{1} \and Adam White\inst{2} \and
Marlos C. Machado\inst{1}  \and Patrick M. Pilarski\inst{1*}}
\authorrunning{Craig Sherstan et al.} 
%
\tocauthor{Craig Sherstan, Adam White, Marlos C. Machado, Patrick M. Pilarski}
\institute{University of Alberta, Edmonton AB, Canada\\
\email{*~pilarski@ualberta.ca}
\and
Indiana University, Bloomington IN, USA}

\maketitle              

\begin{abstract}
Agents of general intelligence deployed in real-world scenarios must adapt to ever-changing environmental conditions. While such adaptive agents may leverage engineered knowledge, they will require the capacity to construct and evaluate knowledge themselves from their own experience in a bottom-up, constructivist fashion. This position paper builds on the idea of encoding knowledge as temporally extended predictions through the use of general value functions. Prior work has focused on learning predictions about externally derived signals about a task or environment (e.g. battery level, joint position). Here we advocate that the agent should also predict internally generated signals regarding its own learning process---for example, an agent's confidence in its learned predictions. Finally, we suggest how such information would be beneficial in creating an introspective agent that is able to learn to make good decisions in a complex, changing world.
%
%
%
%
%
%
%
%
%
%
%
\end{abstract}
%
%
{\bf Predictive Knowledge.} The ability to autonomously construct knowledge directly from experience produced by an agent interacting with the world is a key requirement for general intelligence. One particularly promising form of knowledge that is grounded in experience is \textit{predictive knowledge}---here defined as a collection of multi-step predictions about observable outcomes that are contingent on different ways of behaving. Much like scientific knowledge, predictive knowledge can be maintained and updated by making a prediction, executing a procedure, and observing the outcome and updating the prediction---a process completely independent of human intervention. Experience-grounded predictions are a powerful resource to guide decision making in environments which are too complex or dynamic to be exhaustively anticipated by an engineer \cite{Horde,Modayil2014}.

A {\em value function} from the field of reinforcement learning is one way of representing predictive knowledge. Value functions are a learned or computed mapping from state to the long-term expectation of future reward. Sutton et al. recently introduced a generalization of value functions that makes it possible to specify general predictive questions \cite{Horde}. These {\em general value functions} (GVFs), specify a prediction target as the expected discounted sum of future signals of interest ({\em cumulants}) observed while the agent selects actions according to some decision making policy. 
Temporal discounting is also generalized in GVFs from the conventional exponential weighting of future cumulants to an arbitrary, state-conditional weighting of future cumulants. This enables GVFs to specify a rich class of predictive questions where discounting acts as a stochastic termination function \cite{Modayil2014}. A single GVF specifies a predictive question, and the answer takes the form of an approximate GVF that can be learned by  temporal-difference (TD) learning algorithms solely from unsupervised interaction with the world. A collection of GVFs contributes to the agent's knowledge of the world.  

Ultimately, the purpose of acquiring knowledge is to improve the agent's ability to achieve its goals. The agent's collection of GVFs are only useful to the extent that they help with reward maximization. While GVFs are relatively new, there have been several recent demonstrations of their usefulness in robot tasks, from reflexive action in mobile robots \cite{Modayil2014}, to the control of prosthetic arms \cite{EdwardsAdaptive,SherstanDPCC}. In this paper we take the next step by specifying GVFs whose cumulants are internal signals defined by the agent's own learning process.
%
%
%
%
%
%
%
%
%
%
%
%
%
%
%


{\bf Predicting Internal Signals.} GVFs have been previously used to specify predictions about signals external to the agent---signals in the agent's sensorimotor stream of interactions. However, agents also have access to a set of {\em internal signals} not previously  considered as cumulants for GVFs. Specifically, there are a number of signals available to an agent that relate to the agent's own learning mechanisms---for example, its predictions' errors, weight changes, and other time-varying meta-parameters. There are also a range of signals that quantify the agent's interactions with its sensorimotor stream---for example, statistics about state or feature-space visitation and statistical properties of input signals. Integrating predictions of these internal signals should improve an agent's decision making abilities towards human-level intelligence \cite{Clark2015}.

One representative class of internal signals relates to an agent's certainty in its own predictions. An agent might make better use of its knowledge given some sense of how much each approximate GVF is to be trusted. That is, given a GVF, how \textit{confident} is the agent that the learned prediction is accurate and precise? Methods such as confidence intervals or ensemble forecasting are used in many domains and may also be appropriate here \cite{Wiering2008}. For our purposes, we desire an approach that is compatible with function approximation and supports online and incremental prediction and learning with only linear complexity (in the size of the input features). GVFs and TD methods used to approximate them satisfy these criteria and are therefore a promising approach to incorporating confidence measures. Indeed, this presents an appealing architecture where both predictive questions and measures of their confidence are represented in a single form.

Further, we propose that an agent's decision making process can be improved by using several confidence measures, rather than solely relying on a single value of confidence. Each measure can then provide a different perspective on the accuracy of an approximate GVF, enabling the agent to make more informed decisions. Encoding these measures as GVFs enables these internal predictions to participate in the agent's representation of state \cite{WhiteSurprise}, which can lead to more efficient reward maximization \cite{Rafols2005,Schaul2013} and more accurate prediction \cite{Littman2001}.

\begin{figure}[t!]
\centering
	\begin{tabular}{ccc}
    	\subcaptionbox{Prediction\label{}}{\includegraphics[height=1.3in]{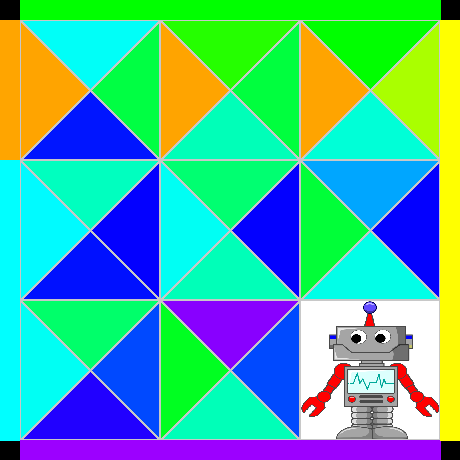}}&
        \subcaptionbox{Visitation\label{}}{\includegraphics[height=1.3in]{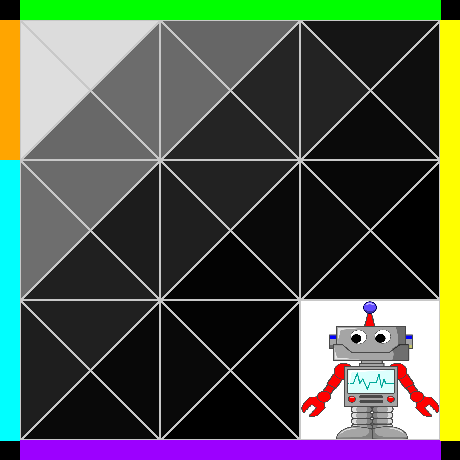}}&      
        \subcaptionbox{Error\label{}}{\includegraphics[height=1.3in]{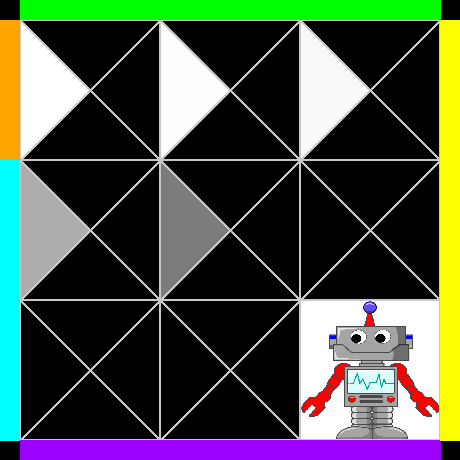}}
    \end{tabular}
    \caption{In this illustrative example, a robot moves through a 3x3 grid world and \textbf{(a)} predicts wall hue (a continuous representation of color in [0,360)) in each of four directions. One GVF is randomly initialized for each direction. A tabular state representation is used, and the robot is only able to observe the hue of a wall when it takes direct action into that wall. At each timestep the hue of the walls may change. In this example, the hue of the upper wall has high variance while all other walls have low or no variance. Each cell of (a) shows the predicted hue in each direction.  \textbf{(b)} The robot moves in a weighted random walk with a preference for moving up and left, as seen in the state-action visitation. High confidence (white) corresponds to high visitation and low confidence (black) corresponds to low visitation (visitation is initialized to 0). \textbf{(c)} The robot makes predictions of the expected squared TD error of the primary hue prediction. High confidence (white) corresponds to low error and low confidence (black) corresponds to high error (error is initialized high). Predictions of variance (not shown) produce a similar pattern to that of (c). The robot can decide to only trust predictions in portions of the world it has visited before, here the upper left. Further, when in the upper-left corner, the robot can see that despite high visitation it should not trust it's upward prediction as error (and variance) remain high. On the other hand, it can trust its leftward prediction as visitation is high and error is low.}
    \label{fig:rainbow}
    
\vspace{-1em}
\end{figure}

While there are many measures that could be useful for an agent in determining confidence \cite{SherstanMSc}, we here provide three examples which are readily represented as GVFs (see Figure~\ref{fig:rainbow} for examples). The first is {\em visitation}. 
A measure of state visitation can be expressed as a GVF by simply using a constant valued cumulant. An agent might reasonably decide to only trust a prediction in states that have been visited many times. The second is {\em prediction error}. On each time step, a temporal-difference error is used to update each approximate GVF \cite{Horde}; this TD error can itself be used as the cumulant of another GVF. Such a prediction gives an agent an expectation of how much each approximate GVF will differ from the true outcome specified by the corresponding GVF.  The third is {\em variance}. The variance of a cumulant or an approximate GVF can easily be represented as the difference between two GVFs, although the process for approximating these nonstationary cumulants is somewhat more involved \cite{WhiteInterval}. 

{\bf Example: Exploration with Confidence Measures.} A collection of approximate GVFs, combined with their corresponding confidence measures provide the agent with a way to measure how much it should trust what it knows. In a safe environment an agent might view low confidence as an opportunity to learn more about its world  \cite{WhiteSurprise,Schmidhuber1991}. In a dangerous environment, low confidence might be a strong indicator to proceed with caution or not at all. Further, confidence itself can be a goal for an agent's behavior. That is, an agent could choose to seek out predictability (high confidence) or novelty (low confidence). This naturally plays a role in the trade off between exploration and exploitation.

%
%
%
%
%
%
%
%
%

\textbf{Concluding remarks}: Predictive knowledge is essential to a generally intelligent agent in maximizing its reward. We advocate that internal measures relating to prediction learning can and should also be represented as GVFs and learned in the same way. These new predictions provide additional knowledge that enables an agent to improve its decision making abilities. GVFs present a novel approach to the general problem of introspection within intelligent agents.
%
%
%
%
%
%


\begin{thebibliography}{14}
%
\bibitem{Horde}
Sutton, R.S., Modayil, J., Delp, M., Degris, T., Pilarski, P.M., White, A., Precup, D.: Horde: A Scalable Real-time Architecture for Learning Knowledge from Unsupervised Sensorimotor Interaction Categories and Subject Descriptors. In: Int. Conf. on Autonomous Agents and Multi-Agent Systems, pp. 761--768 (2011).

\bibitem{Modayil2014}
Modayil, J., White, A., Sutton, R.S.: Multi-Timescale Nexting in a Reinforcement Learning Robot. Adapt. Behav. 22, pp. 146--160 (2014).

\bibitem{EdwardsAdaptive}
Edwards, A.L., Dawson, M.R., Hebert, J.S., Sherstan, C., Sutton, R.S., Chan, K.M., Pilarski, P.M.: Application of real-time machine learning to myoelectric prosthesis control: A case series in adaptive switching. Prosthet. Orthot. Int., published online ahead of print, pp. 1--9 (2015).

\bibitem{SherstanDPCC}
Sherstan, C., Modayil, J., Pilarski, P.M.: A Collaborative Approach to the Simultaneous Multi-joint Control of a Prosthetic Arm. In: Int. Conf. on Rehabilitation Robotics, pp. 13--18, Singapore, Singapore (2015).

\bibitem{Clark2015}
Clark, A.: Surfing Uncertainty: Prediction, Action, and the Embodied Mind. Oxford University Press (2015).

\bibitem{Wiering2008}
Wiering, M.A., van Hasselt, H.: Ensemble algorithms in reinforcement learning. IEEE Trans. Syst. Man, Cybern. Part B Cybern. 38, 4, pp. 930--936 (2008).

\bibitem{WhiteSurprise}
 White, A.: Developing a predictive approach to knowledge. PhD Thesis. University of Alberta (2015).


 \bibitem{Rafols2005} 
Rafols, E.J., Ring, M.B., Sutton, R.S., Tanner, B.: Using predictive representations to improve generalization in reinforcement learning. In: Int. Joint Conf. on Artificial Intelligence, pp. 835--840 (2005).

\bibitem{Schaul2013}
Schaul, T., Ring, M.: Better Generalization with Forecasts. In: Int. Joint Conf. on Artificial Intelligence, pp. 1656--1662, Beijing, China (2013).

\bibitem{Littman2001}
Littman, M.L., Sutton, R.S., Singh, S.: Predictive Representations of State. Advances in Neural Information Processing Systems 14, pp. 1555--1561 (2001).

\bibitem{SherstanMSc}
Sherstan, C.: Towards Prosthetic Arms as Wearable Intelligent Robots. MSc Thesis. University of Alberta (2015).


\bibitem{WhiteInterval}
White, M., White, A.: Interval Estimation for Reinforcement-Learning Algorithms in Continuous-State Domains. In: Advances in Neural Information Processing Systems 23, pp. 2433--2441 (2010).


 





\bibitem{Schmidhuber1991}
Schmidhuber, J.: Curious model-building control systems. In: IEEE Int. Joint Conf. on Neural Networks. pp. 1458--1463, Singapore, Singapore (1991).

\end{thebibliography}
\end{document}